\documentclass[runningheads]{llncs}

\usepackage{eccv}

\usepackage{eccvabbrv}

\usepackage{graphicx}
\usepackage{booktabs}
\usepackage{adjustbox}
\usepackage{multirow}
\usepackage[table,xcdraw]{xcolor}
\usepackage{colortbl}
\usepackage{graphicx}
\usepackage{array}
\usepackage[accsupp]{axessibility}  

\usepackage{hyperref}

\usepackage{orcidlink}

\begin{document}

\title{Multi-modality Image Fusion under Adverse Weather: Mask-Guided Feature Restoration and Interaction} 
\titlerunning{Multi-modality Image Fusion under Adverse Weather}

\author{
Xilai Li$^{1}$ \and
Xiaosong Li$^{1,\dagger}$ \and
Haishu Tan$^{1}$ \and
Tao Ye$^{2}$ \and
Huafeng Li$^{3}$ \and
Hongbin Wang$^{3}$
}

\authorrunning{X. Li et al.}

\institute{
Foshan University, Foshan, China\\
\email{20210300236@stu.fosu.edu.cn}, \email{lixiaosong@buaa.edu.cn}
\and
China University of Mining and Technology, Beijing, China
\and
Kunming University of Science and Technology, Kunming, China
}

\begingroup
\renewcommand{\thefootnote}{\ensuremath{\dagger}}
\footnotetext{Corresponding author.}
\endgroup

\maketitle

\begin{abstract}
Multi-modality image fusion (MMIF) enhances scene representation by exploiting complementary cues from different modalities. Adverse weather, however, causes significant image degradation, disrupting feature representation and requiring simultaneous feature restoration and cross-modal complementarity. Existing methods often struggle with effective representation learning under such conditions, limiting their practical performance. To address these challenges, we propose a mask-guided MMIF method that integrates feature restoration and interaction. We first introduce "Pseudo Ground Truth" to simplify training, promoting faster and more effective feature learning. Then, we design a mask generation mechanism based on the mapping relationship between the fused result and the source images, quantifying the relative contribution of each modality during the fusion process. By incorporating the proposed mask-guided cross-modal cross-attention mechanism, the network is encouraged to selectively attend to informative features during modality interaction, mitigating the risk of overfitting to the static distribution of the "Pseudo Ground Truth". Additionally, we propose a mask-guided learning strategy and a task-coupled degradation-aware learning strategy to balance feature restoration and interaction. Extensive experiments on synthetic and real-world datasets demonstrate that our method surpasses state-of-the-art approaches in visual quality, quantitative metrics, and downstream tasks. The source code is available at \href{https://github.com/ixilai/AMG-Fuse}{https://github.com/ixilai/AMG-Fuse}.
  \keywords{Image fusion \and Adverse weather \and Feature restoration \and Feature interaction}
\end{abstract}

\section{Introduction}
Multi-modality image fusion (MMIF) \cite{r124,r1,r3,r79,r121,r131} combines complementary information from multiple modalities to produce richer scene representations. For instance, in infrared and visible image fusion (IVIF), visible images capture texture details critical for semantic interpretation, while infrared images highlight salient targets via thermal radiation \cite{r24}. By integrating these modalities, IVIF enhances downstream tasks like object detection \cite{r6}, semantic segmentation \cite{r5}, and depth estimation \cite{r106}.

Existing studies \cite{r18,r77,r107} have extensively investigated IVIF architectures, mainly focusing on ideal scenarios. Given a pair of registered infrared ($IR$) and visible ($VI$) images, the fusion task typically generates fused outputs by minimizing an L1-norm-based loss function. The optimization objective for conventional IVIF can be expressed as:
\begin{equation}
  \theta^* = \arg\min_{\theta} \left( \lambda_1 \left\| f({VI}, {IR}; \theta) - {VI} \right\|_1 + \lambda_2 \left\| f({VI}, {IR}; \theta) - {IR} \right\|_1 \right)
  \label{eq1}
\end{equation}
Here, $f(\cdot)$ represents the fusion network, $\theta$ denotes the network parameters, and $\lambda_1$, $\lambda_2$ are weights balancing the contributions of both modalities. This objective enables the network to extract meaningful information from multiple modalities effectively. Some methods \cite{r24,r18} augment this with additional loss functions to further optimize fusion performance.

However, real-world complexities like noise \cite{r92,r108}, adverse weather \cite{r91}, and motion blur \cite{r109} pose significant challenges to IVIF adaptability. Recent works \cite{r91,r105,r89} addressing these conditions can be categorized into two types. The first follows a “restoration + fusion” paradigm, where each modality is first processed by a restoration network before being input into a fusion network. Nevertheless, this two-stage design introduces several issues: (1) Restoration focuses on intra-modal recovery, while fusion targets inter-modal complementarity, leading to unstable optimization.  (2) Error accumulation: Artifacts from the restoration stage may propagate into the fusion process, amplifying degraded features and impairing reconstruction.

\begin{figure}[t]
  \centering
   \includegraphics[width=0.9\linewidth]{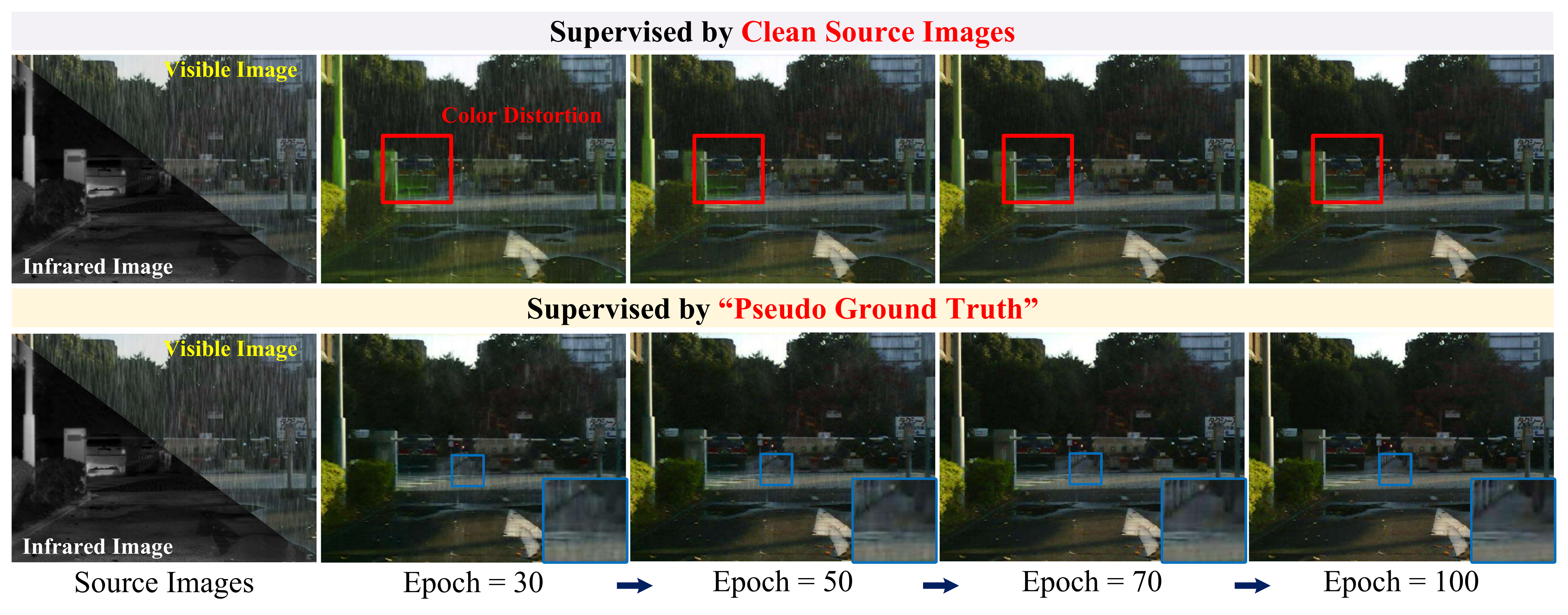}
   \caption{Comparison of optimization behavior under clean-source-image and "Pseudo Ground Truth" supervision.}
   \label{fig1}
\end{figure}
To address these issues, some studies \cite{r110,r111,r118,r127,r128} adopt a second strategy by using "Pseudo Ground Truth" as supervision. Specifically, the "Pseudo Ground Truth" refers to fusion results obtained by existing methods using clean $VI$ and $IR$ images as supervision targets. The optimization objective of this second strategy can be formulated as follows:
\begin{equation}
\theta^* = \arg\min_{\theta} \left\| f({VI}, {IR}; \theta) - {GT}_{{{\mathrm{Pse}}}} \right\|_1
  \label{eq2}
\end{equation}
where ${GT}_{{\mathrm{Pse}}}$ denotes the "Pseudo Ground Truth". As shown in Figure~\ref{fig1}, source-image supervision makes the network struggle with joint feature extraction and degradation removal, often leaving rain streaks and color distortions. In contrast, ``Pseudo Ground Truth'' supervision simplifies optimization and helps preserve global structures and fine details.

\begin{figure}[t]
  \centering
   \includegraphics[width=0.85\linewidth]{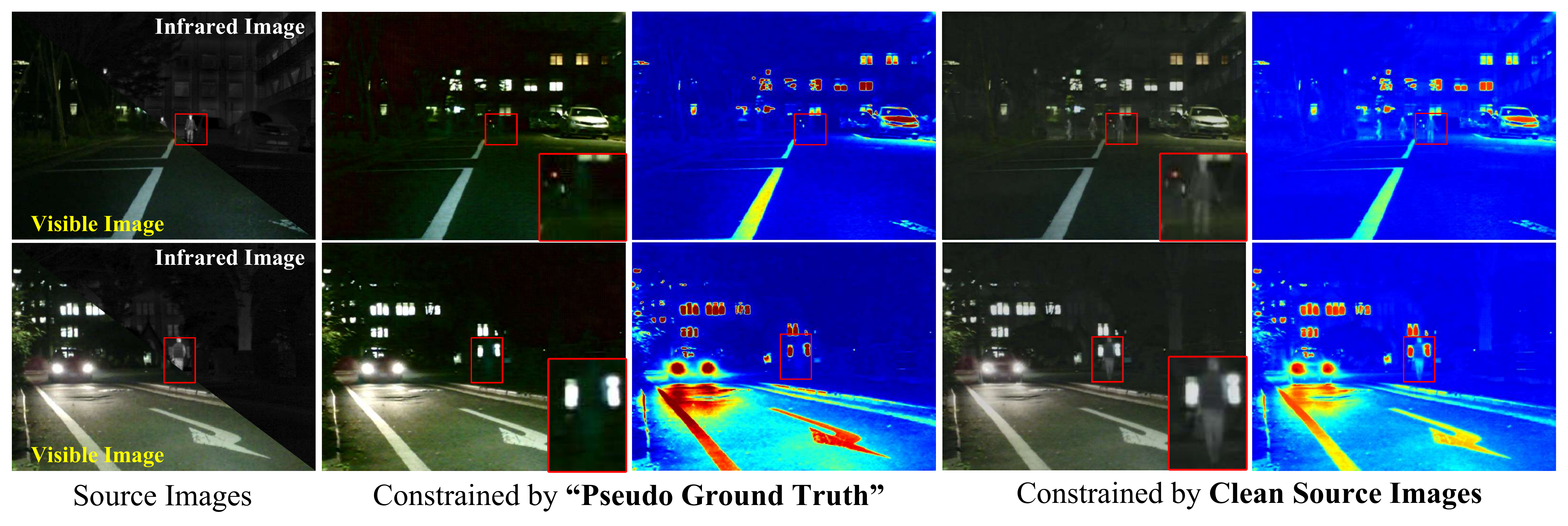}
   \caption{Visualization of modality bias caused by "Pseudo Ground Truth" supervision in clean scenes.}
   \label{fig2}
\end{figure}

However, although this approach reduces optimization complexity, it cannot fully resolve fusion challenges in complex scenarios, since ``Pseudo Ground Truth'' generated from clean images by existing methods still carries information loss and modality bias.
Moreover, networks trained with ``Pseudo Ground Truth'' may overlook complementary multi-modal cues, weakening generalization. As shown in Figure~\ref{fig2}, such models underutilize infrared thermal information in clean scenes, causing the loss of critical target regions. This highlights a key limitation: the strategy fits static distributions of pseudo targets rather than learning the dynamic mechanisms of IVIF.
This raises a crucial question: \emph{Can we harness the advantages of "Pseudo Ground Truth" while enabling the network to dynamically mine cross-modal information and remain degradation-aware?}

To address these issues, we propose a mask-guided adverse-weather image fusion framework (AMG-Fuse). We first analyze the relationship between source images and fusion results and derive a mask representation to decouple modality-specific components in the fused image. This mask separates multi-modal features in the ``Pseudo Ground Truth'' and serves as an external prior for explicit cross-modal interaction modeling during training, encouraging the network to learn a dynamic fusion process. We further design a Mask-Guided Feature Extraction Module (MFEM) that uses the mask to restore degraded features and enhance cross-modal interaction.
During training, the proposed Mask-Guided Learning Strategy (MGLS) and Task-Coupled Degradation-Aware Learning Strategy (TDAS) collaboratively supervise the fusion network, facilitating the learning of clearer and more complementary feature representations. Our main contributions are as follows: 

\begin{itemize}

\item We propose a unified multi-modality image fusion method for complex scenarios, enabling simultaneous feature restoration and modality interaction within a unified network. 

\item We propose a \textbf{mask-guided feature extraction module} that effectively reconstructs the fusion features. Additionally, we design a mask-guided network learning strategy to address the lack of dynamic cross-modal interaction modeling.

\item We propose a \textbf{task-coupled degradation-aware learning strategy} that leverages the restoration task as a meaningful supervisory signal to enhance degradation perception and guide the feature fusion process.

\item We conduct extensive experiments under three adverse weather conditions to validate the effectiveness of the proposed method, and demonstrate its potential for real-world applications through downstream tasks.
\end{itemize}

\section{Related Works}
\subsection{Learning based IVIF Methods}
Recent advances in deep learning have greatly promoted IVIF algorithms \cite{r112,r113,r4}. Existing studies under ideal conditions mainly focus on cross-modal information interaction \cite{r112,r24,r18,r64}, unified fusion framework design \cite{r19,r114,r63,r115}, and task-driven optimization \cite{r21,r17,r116,r117,r122}. Cross-modal interaction typically uses feature extractors, such as convolution or self-attention, to distinguish shared and modality-specific cues for effective fusion. For example, Zhao et al. \cite{r24} designed a correlation-driven dual-branch Transformer-CNN to separate global and local features and preserve salient details. With larger datasets and stronger computation, unified and downstream-oriented fusion frameworks have received increasing attention, such as Wu et al. \cite{r116} distilling the Segment Anything Model into the fusion network.

\subsection{IVIF Methods for Complex Scenes}
The complexity of real-world environments presents great challenges for IVIF algorithms in practical applications. Beyond fusion tasks in ideal conditions, researchers \cite{r12,r91,r129,r130,r123} have shifted their focus to image fusion in complex conditions such as noise, overexposure, and adverse weather. The primary challenge in complex scenes is not only achieving cross-modal interaction, but also accurately distinguishing useful features from degradations and ensuring effective restoration. For example, Yi et al. \cite{r65} proposed a degradation-aware interactive fusion algorithm guided by semantic text, using textual information as an external prior to help the network focus on existing degradations. Li et al. \cite{r91} combined physical models and introduced an all-weather IVIF algorithm, incorporating various priors into the student model through distillation learning. Although the above methods \cite{r65,r89,r91} have advanced fusion research in complex scenarios, they mainly focus on feature restoration and do not sufficiently explore the paradigm gap between complex and ideal fusion.

\begin{figure}[t]
  \centering
   \includegraphics[width=1\linewidth]{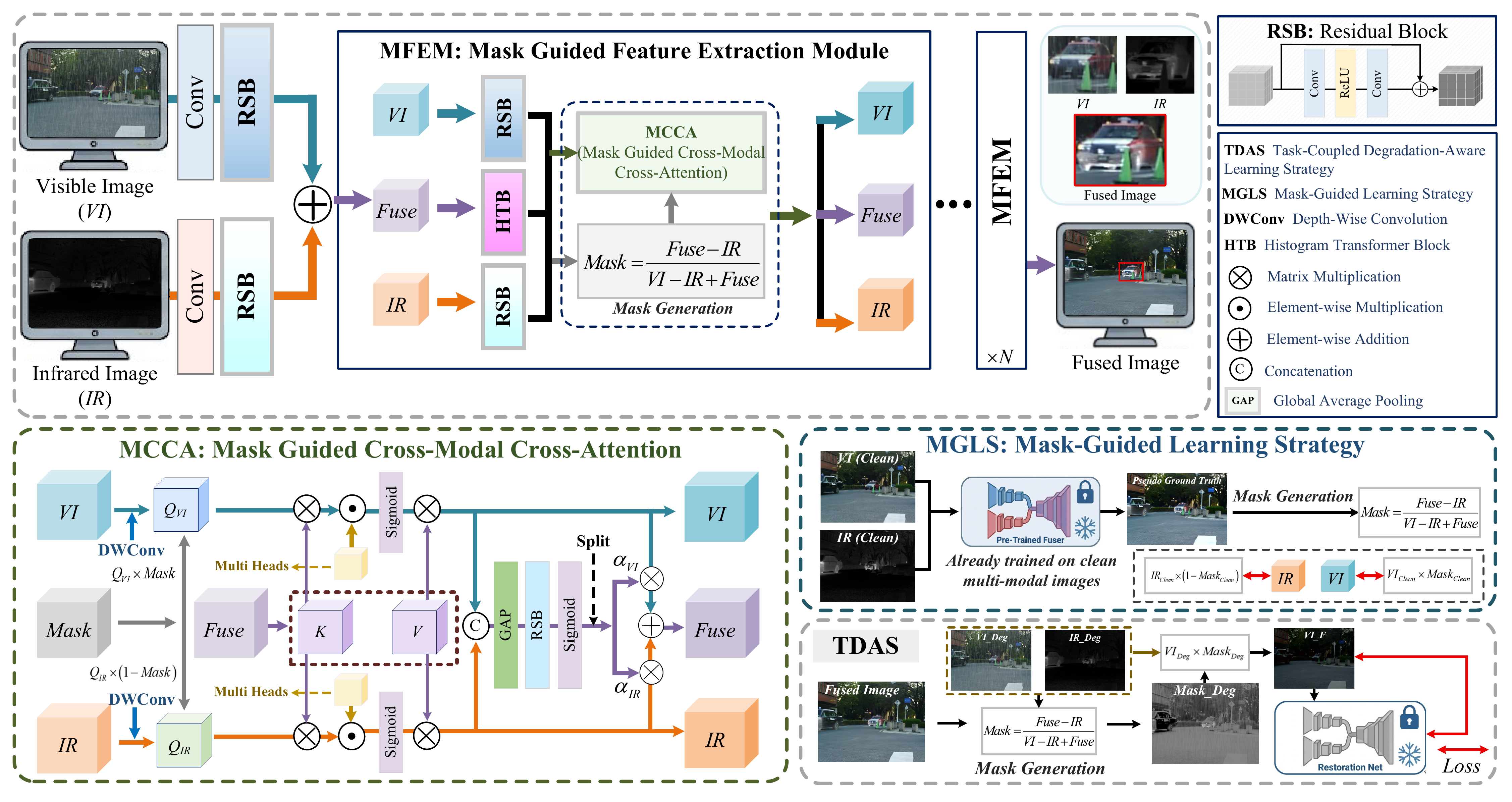}
   \caption{The flowchart of the proposed method.}
   \label{fig3}
\end{figure}

\section{Proposed Method}
The workflow of the proposed algorithm is illustrated in Figure~\ref{fig3}. First, convolutional and residual blocks are employed to enhance the feature representation of the input image. The extracted multi-modal features are then combined to generate initial fused features. Subsequently, residual blocks further refine modality-specific features, while the fused features are processed by the Histogram Transformer Block (HTB) \cite{r119} to extract prominent information. HTB segments spatial features according to pixel intensity and applies self-attention to different intensity ranges, enabling it to capture similar degradation patterns across long spatial distances \cite{r119}. Finally, the multi-modal and fused features are jointly refined by the Mask-Guided Cross-Modal Cross-Attention (MCCA) module to produce the final fused image.

\subsection{Mask Construction from Fusion Formulation}
\begin{figure}[t]
  \centering
   \includegraphics[width=0.7\linewidth]{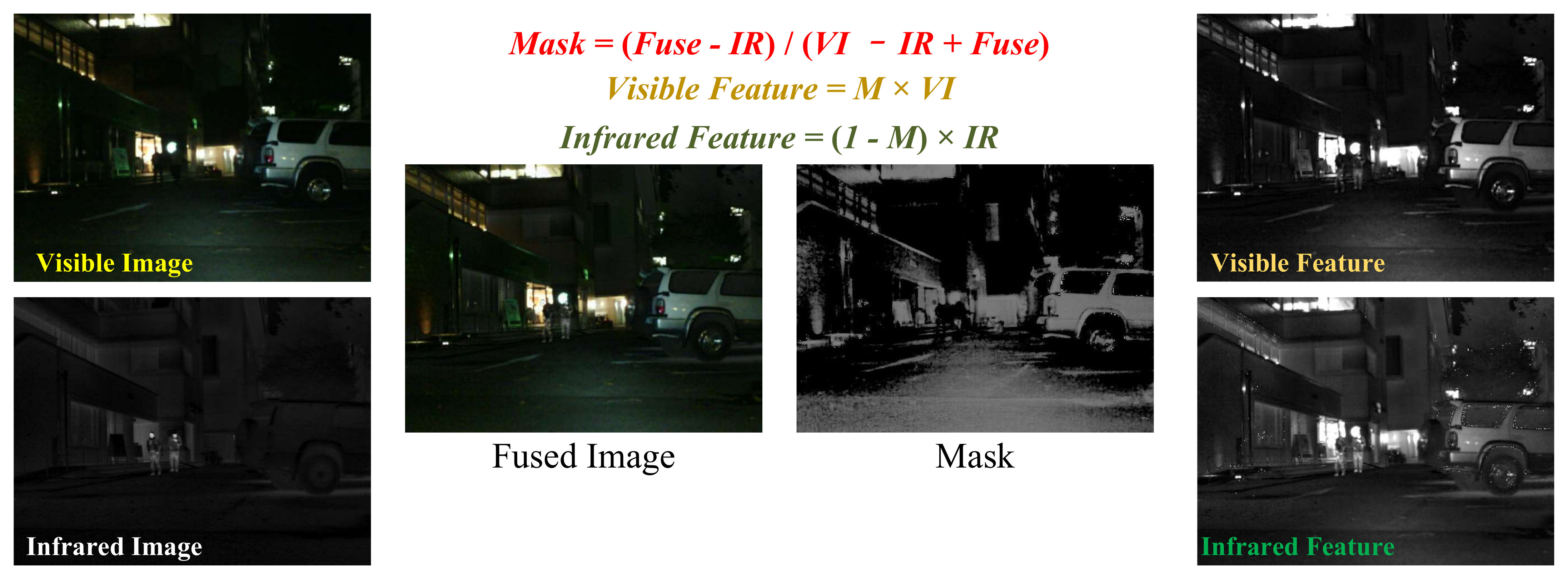}
   \caption{Visualization of Modality-Contribution Mask and Feature Decomposition.}
   \label{fig4}
\end{figure}

Although the introduction of "Pseudo Ground Truth" can ease the training process, it may also lead the network to directly replicate its features rather than learning the underlying modality allocation knowledge it conveys. To overcome this limitation, we decouple it by introducing the Mask.
The core objective of the IVIF task is to suppress cross-modal redundancy and effectively extract complementary information. Disregarding feature enhancement and reduction, the fusion result ($Fuse$) can be expressed as:
\begin{equation}
{Fuse} = M \times {VI} + (1 - M) \times {IR} + \varepsilon
  \label{eq3}
\end{equation}
Here, $M$ denotes the mask for modality-specific weight allocation, and $\varepsilon$ represents the error and noise term. Given that ${Fuse}$, ${VI}$, and ${IR}$ are known, the expression for $M$ can be derived as:
\begin{equation}
M = \frac{{Fuse} - {IR}}{{VI} - {IR}} + \varepsilon_M
\label{eq4}
\end{equation}
In real-world scenarios, directly subtracting infrared images from visible images can lead to misleading or unstable behaviour. Under haze or snow, visible images often exhibit excessive or locally overexposed brightness, whereas infrared images show reduced contrast and lower pixel values. After network optimisation and degradation removal, the fused output enhances the structural and target information from the infrared component, resulting in a more stable ($Fuse - IR$) distribution that better reflects the true scene brightness. However, because the brightness bias in the denominator of Equation~\ref{eq4} is driven by degradation factors from the visible image rather than semantic content, degraded information dominates the weighting, violating the decoupling principle of mask and preventing it from capturing the actual modal distribution. Conversely, in night-time scenes, visible images contain minimal effective texture due to insufficient illumination, while infrared images remain relatively stable and may present stronger brightness cues. This causes ($VI - IR$) to be negative or near zero over large regions, making Equation~\ref{eq4} numerically unstable when generating the mask. To address this issue, we rewrite the expression of $M$ as follows:
\begin{equation}
M = \frac{{Fuse} - {IR}}{{VI} - {IR} + {Fuse}} + \varepsilon_M
\label{eq5}
\end{equation}
Incorporating the $Fuse$ information into the denominator effectively prevents the brightness bias of the visible image from erroneously amplifying the mask and avoids extreme numerical instability in the denominator. Finally, with $M$, we can then access the representation of different fusion algorithms for different modal information. As illustrated in Figure~\ref{fig4}, we visualize the constructed mask and the decoupled multi-modal features. By guiding the network to learn the distribution pattern of multi-modal information in the "Pseudo Ground Truth", our method dynamically models the contribution of each modality, avoiding superficial fitting to the overall scene.

\subsection{Mask Guided Feature Extraction Module}
Training directly on pseudo targets can lead the network to replicate their inherent biases, emphasizing superficial feature reproduction over effective cross-modal interactions. To address this, we propose the Mask-Guided Feature Extraction Module (MFEM).

The MFEM is designed based on Equation~\ref{eq3} and Equation~\ref{eq5}.  Firstly, multi-modality features are fed into the Residual Block (RSB), while the fused features are passed to the Histogram Transformer Block (HTB) for optimization. The resulting outputs are then used to compute the mask $M$. The MCCA module is proposed to enable the network to learn the multi-modal interaction patterns of the "Pseudo Ground Truth". The MCCA employs a cross-attention mechanism, where multi-modal features serve as queries guided by the mask, while the fused features act as keys and values. 
To improve local information modeling, depthwise separable convolutions are introduced in both the query and key-value branches to expand the local receptive field. The mask is then used to weight the queries $Q_{VI}$ and $Q_{IR}$, guiding the network to focus on the significant features in each modality. This enables  decoupling and recombination of multi-modal features within the fusion space.

\subsection{Mask Guided Learning Strategy}
To better leverage "Pseudo Ground Truth" for learning multi-modal interaction features, we propose a Mask Guided Learning Strategy (MGLS). Specifically, the mask $M_{\mathrm{Pse}}$ is computed based on the "Pseudo Ground Truth" ($GT_{\mathrm{Pse}}$) and the clean multi-modality source images $VI_{\mathrm{C}}$ and $IR_{\mathrm{C}}$, as shown in Equation~\ref{eq5}. This mask is then used to decouple the multi-modality features $F_{VI}$ and $F_{IR}$ in the "Pseudo Ground Truth".
\begin{equation}
F_{VI} = M_{\mathrm{Pse}} \times VI_{\mathrm{C}}, \quad F_{IR} = (1 - M_{\mathrm{Pse}}) \times IR_{\mathrm{C}}
\label{eq6}
\end{equation}
Here, \( F_{VI} \) and \( F_{IR} \) are the modality allocation maps in \( GT_{\mathrm{Pse}} \). Then, we compute the L1 loss for the corresponding multi-modal features to constrain the network in learning the multi-modal information distribution pattern in the "Pseudo Ground Truth". Given the network output the multi-modal features as \( \hat{F}_{VI} \) and \( \hat{F}_{IR} \), the proposed MGLS loss (\( \mathcal{L}_{MGLS} \)) is computed as follows:
\begin{equation}
\mathcal{L}_{MGLS} = \frac{1}{HW} \left( \| \hat{F}_{VI} - F_{VI} \|_1 + \| \hat{F}_{IR} - F_{IR} \|_1 \right)
\label{eq7}
\end{equation}
where $H$ and $W$ represent the height and width of the image respectively.

\subsection{Task-Coupled Degradation-Aware Learning Strategy}
\begin{figure}[t]
  \centering
   \includegraphics[width=0.9\linewidth]{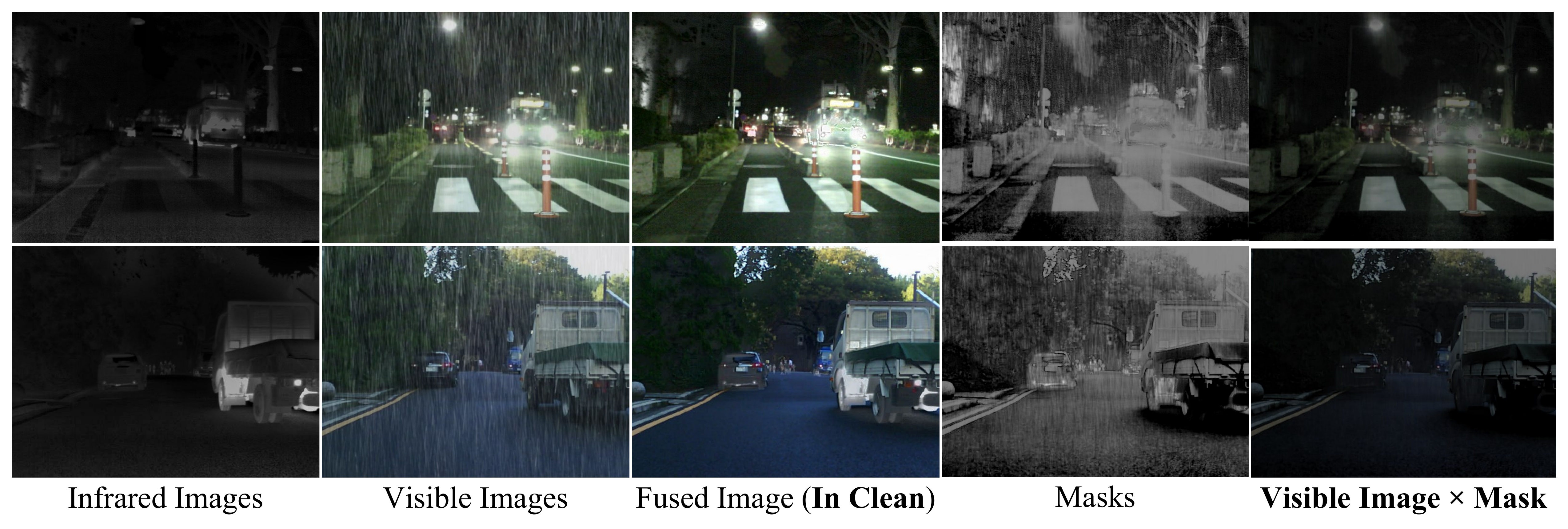}
   \caption{Visualization of Mask-Guided Degradation Suppression in Visible Feature Reweighting.}
   \label{fig5}
\end{figure}

Equation~\ref{eq3} defines the synthesis paradigm of conventional image fusion. Therefore, after determining the mask, the allocation strategies of the existing methods for different modal information can be indirectly obtained. In complex scenes, the two input images ($VI_{Deg}$ and $IR_{Deg}$) are often degraded. At this point, the goal of the fusion network is to output a clear fusion result that contains complementary multi-modal features. This process can be modeled analogously using Equation~\ref{eq3}.
\begin{equation}
Fuse = M_{\mathrm{Deg}} \times VI_{Deg} + (1 - M_{\mathrm{Deg}}) \times IR_{Deg} + \varepsilon
\label{eq8}
\end{equation}
Since $Fuse$ is a non-degraded image, the visible modal information obtained by $M_{\mathrm{Deg}} \times VI_{Deg}$ in Equation~\ref{eq8} should be degradation-free. Thus, the mask can effectively capture the distribution of the degraded area. This is demonstrated in Figure~\ref{fig5}, which shows that most of the rain streaks have been successfully suppressed or removed.
Based on the above observation results, we propose the Task-Coupled Degradation-Aware Learning Strategy (TDAS), aiming to guide the fusion network to prioritize processing clearer and more prominent regions. For the restoration task, when the input image is degradation-free, the restoration network tends to generate an output that closely resembles the original image. Therefore, we define the TDAS loss as follows:

\begin{equation}
\mathcal{L}_{\mathrm{TDAS}} = \frac{1}{H \times W} \left\| {VI}_F - \mathcal{R}({VI}_F) \right\|_1
\label{eq9}
\end{equation}
where $\mathcal{R}(\cdot)$ represents the restoration model \cite{r66}, and ${VI}_F= M_{\mathrm{Deg}} \times VI_{Deg}$. When ${VI}_F$ is similar enough to the output $\mathcal{R}({VI}_F)$ of the restoration network, it means that the fusion network has effectively restored the clear features.

\subsection{Loss Functions}
During training, in addition to MGLS and TDAS, source image supervision is further introduced to enhance the ability of the model to capture multi-modal distributions by imposing constraints on the fusion output. Specifically, we adopt a gradient loss and a color consistency loss to preserve structural details and color distribution consistency, respectively. The gradient loss is defined as follows:
\begin{equation}
\mathcal{L}_{\text{grad}} = \frac{1}{H \times W} \left\| \nabla {Fuse} - \max\left( |\nabla {VI}_{C}|, |\nabla {IR}_{C}| \right) \right\|_1
\label{eq10}
\end{equation}
The color consistency loss \cite{r65} is expressed as follows,
\begin{equation}
\mathcal{L}_{\text{color}} = \frac{1}{H \times W} \left\| F_{\text{CbCr}}({Fuse}) - F_{\text{CbCr}}({VI}_{C}) \right\|_1
\end{equation}
where $F_{\text{CbCr}}$ represents the function of transferring the image space to the CbCr color space. The total loss function is expressed as follows,
\begin{equation}
\mathcal{L}_{\text{total}} = \lambda \times \mathcal{L}_{\text{MGLS}} + \mathcal{L}_{\text{TDAS}} + \mathcal{L}_{\text{color}} + \mathcal{L}_{\text{grad}}
\end{equation}
where $\lambda$ is the decay coefficient, which gradually decreases as the number of training epochs increases. All loss terms except $\mathcal{L}_{\text{MGLS}}$ are assigned a weight of 1, ensuring balanced contributions from gradient and color consistency constraints without requiring careful hyperparameter tuning.

\section{Experiments}
\subsection{Experimental Setups}

\textbf{Datasets:}
Our experiments are conducted in two scenarios: adverse weather and ideal environments. For the adverse weather experiments, we selected 1,000 images from each of the three distinct weather conditions (Snow, Rain, Haze) in the AWMM-100k dataset \cite{r91} for training, and 150 images for testing. In addition, we evaluated our method on real degraded images from AWMM-100k to further validate its performance under authentic degradation. For the ideal environment experiments, we utilized three public datasets: M3FD \cite{r17}, MSRS \cite{r62}, and LLVIP \cite{r76}.

\textbf{Comparative Methods:}
We adopted the "restoration + fusion" scheme for performance comparison. The adaptive image restoration network (AdaIR) \cite{r120} was employed as the restoration module and retrained on the AWMM-100k dataset. The fusion methods include LRRNet \cite{r32}, Text-DiFuse \cite{r105}, EMMA \cite{r77}, Text-IF \cite{r65}, GIFNet \cite{r114}, SAGE \cite{r116}, and AWFusion \cite{r91}. Among these, only Text-DiFuse, Text-IF and AWFusion are specifically designed for complex scenes. However, as Text-DiFuse and Text-IF lack the capability to handle adverse weather conditions, they are categorized as standard fusion methods. 

\textbf{Metrics:}
We evaluated the performance of different image fusion methods from multiple perspectives using five types of metrics: (1) Information Theory-Based: Normalized Mutual Information ($Q_{MI}$); (2) Feature-Based: Gradient-Based Fusion Performance ($Q_{G}$) and Multiscale Image Fusion Metric ($Q_{M}$); (3) Structural Similarity-Based: Structural Similarity Index Measure ($SSIM$); (4) Human Perception Inspired: Chen-Blum Metric ($Q_{CB}$) and Visual Information Fidelity ($VIF$); and (5) Other Comprehensive Metrics: Normalized Weighted Performance Metric ($Q^{AB/F}$) \cite{r69}. For all metrics, a higher value indicates better image quality.

\textbf{Training details:}
During the training process, we randomly cropped images from the AWMM-100k training set into $168 \times 168$ patches and trained the model for 200 epochs. The Adam optimizer was used with an initial learning rate of $1 \times 10^{-3}$ and a batch size of 2. All the experiments were conducted in the PyTorch 2.1.1 environment, and the server was equipped with four GeForce RTX 3090 GPUs. 

\begin{figure}[t]
  \centering
   \includegraphics[width=1\linewidth]{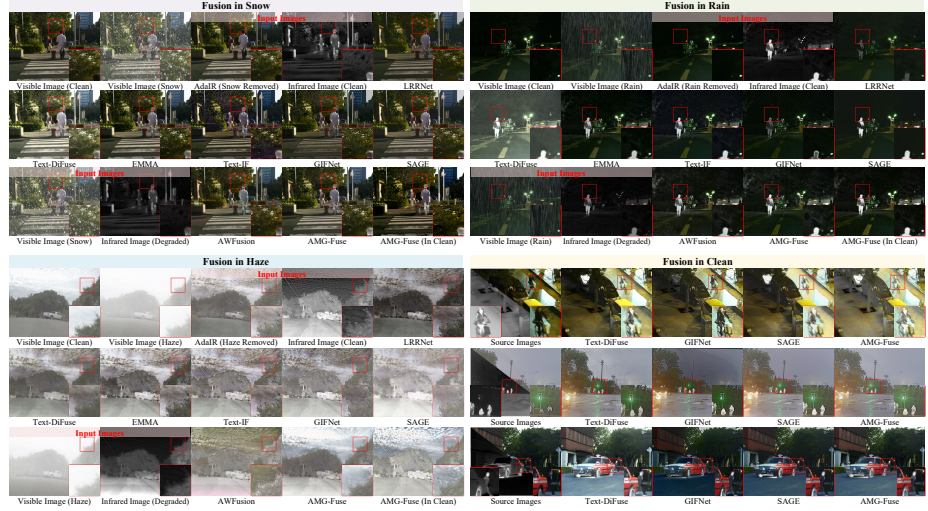}
   \caption{Qualitative comparison results of all methods in the four scenarios.}
   \label{fig6}
\end{figure}

\definecolor{bestred}{HTML}{FADADE}
\definecolor{secondgreen}{HTML}{E3F2D9}

\begin{table}[t]
\centering
\caption{Quantitative comparison of all methods in three adverse weather scenarios. The marked \colorbox{bestred}{red} indicates the best score and the marked \colorbox{secondgreen}{green} indicates the second-best score.}
\label{tab1}
\renewcommand{\arraystretch}{1}
\setlength{\tabcolsep}{4pt}     
\begin{adjustbox}{width=\textwidth}
\begin{tabular}{c|l|c|c|ccccccc}
\hline
\textbf{Source} & \textbf{Methods} & \textbf{Pub.} & \textbf{Restoration} & $Q_{MI} \uparrow$ & $Q_{G} \uparrow$ & $Q_{M} \uparrow$ & $VIF \uparrow$ & $Q_{CB} \uparrow$ & $SSIM \uparrow$ & $Q^{AB/F} \uparrow$ \\ \hline
\multirow{8}{*}{Snow} & LRRNet & PAMI 23 & \multirow{6}{*}{AdaIR} & 0.4403 & 0.2849 & 0.3779 & 0.2588 & 0.3661 & 0.1566 & 0.3970 \\
 & Text-DiFuse & NeurIPS 24 &  & 0.4087 & 0.3047 & 0.3633 & 0.2433 & 0.4264 & 0.2821 & 0.4130 \\
 & EMMA & CVPR 24 &  & \cellcolor{bestred}0.5335 & \cellcolor{secondgreen}0.4235 & 0.5353 & 0.3252 & \cellcolor{secondgreen}0.5011 & 0.4096 & \cellcolor{secondgreen}0.5470 \\
 & Text-IF & CVPR 24 &  & 0.4399 & 0.4199 & \cellcolor{secondgreen}0.5942 & 0.3300 & 0.4929 & \cellcolor{secondgreen}0.4298 & 0.5419 \\
 & GIFNet & CVPR 25 &  & 0.3388 & 0.3282 & 0.3305 & 0.2732 & 0.4854 & 0.3722 & 0.3882 \\
 & SAGE & CVPR 25 &  & 0.4693 & 0.3149 & 0.4752 & \cellcolor{bestred}0.3819 & 0.4421 & 0.3474 & 0.4585 \\ \cline{2-11} 
 & AWFusion & INFFus 26 & \multirow{2}{*}{$\times$} & 0.4668 & 0.3552 & 0.5535 & 0.3024 & 0.4798 & 0.3734 & 0.4937 \\
 & \textbf{AMG-Fuse} & \textbf{--} &  & \cellcolor{secondgreen}0.4960 & \cellcolor{bestred}0.4240 & \cellcolor{bestred}0.6408 & \cellcolor{secondgreen}0.3463 & \cellcolor{bestred}0.5062 & \cellcolor{bestred}0.4302 & \cellcolor{bestred}0.5519 \\ \hline
 \hline
\multirow{8}{*}{Rain} & LRRNet & PAMI 23 & \multirow{6}{*}{AdaIR} & 0.4002 & 0.2306 & 0.4119 & 0.2287 & 0.3335 & 0.0453 & 0.3334 \\
 & Text-DiFuse & NeurIPS 24 &  & 0.3546 & 0.2442 & 0.3645 & 0.2320 & 0.3909 & 0.1962 & 0.3329 \\
 & EMMA & CVPR 24 &  & \cellcolor{bestred}0.4536 & 0.3745 & 0.5871 & 0.3384 & 0.4785 & 0.3700 & 0.4866 \\
 & Text-IF & CVPR 24 &  & 0.3717 & \cellcolor{secondgreen}0.4085 & 0.6375 & \cellcolor{secondgreen}0.3489 & 0.4824 & \cellcolor{secondgreen}0.4046 & \cellcolor{secondgreen}0.5149 \\
 & GIFNet & CVPR 25 &  & 0.2886 & 0.2975 & 0.3686 & 0.3091 & \cellcolor{secondgreen}0.4834 & 0.3474 & 0.3544 \\
 & SAGE & CVPR 25 &  & 0.3994 & 0.2286 & 0.5086 & 0.3478 & 0.3964 & 0.2734 & 0.3670 \\ \cline{2-11} 
 & AWFusion & INFFus 26 & \multirow{2}{*}{$\times$} & 0.3509 & 0.3334 & \cellcolor{secondgreen}0.6428 & 0.2997 & 0.4691 & 0.3528 & 0.4618 \\
 & \textbf{AMG-Fuse} & \textbf{--} &  & \cellcolor{secondgreen}0.4084 & \cellcolor{bestred}0.4163 & \cellcolor{bestred}0.6940 & \cellcolor{bestred}0.3582 & \cellcolor{bestred}0.4879 & \cellcolor{bestred}0.4080 & \cellcolor{bestred}0.5184 \\ \hline
 \hline
\multirow{8}{*}{Haze} & LRRNet & PAMI 23 & \multirow{6}{*}{AdaIR} & 0.2411 & 0.3044 & 0.3531 & 0.2496 & 0.4306 & 0.2031 & 0.3701 \\
 & Text-DiFuse & NeurIPS 24 &  & 0.2304 & 0.3137 & 0.3463 & 0.2234 & 0.4391 & 0.3035 & 0.4021 \\
 & EMMA & CVPR 24 &  & 0.2345 & 0.3553 & 0.4149 & 0.2559 & 0.4567 & 0.3296 & 0.4617 \\
 & Text-IF & CVPR 24 &  & 0.2312 & 0.3999 & 0.4418 & 0.2779 & 0.4646 & 0.3443 & 0.4883 \\
 & GIFNet & CVPR 25 &  & 0.2451 & 0.3124 & 0.3181 & 0.2401 & \cellcolor{secondgreen}0.4784 & 0.3463 & 0.3478 \\
 & SAGE & CVPR 25 &  & 0.2312 & 0.3416 & 0.3948 & 0.2850 & 0.4553 & 0.3316 & 0.4336 \\ \cline{2-11} 
 & AWFusion & INFFus 26 & \multirow{2}{*}{$\times$} & \cellcolor{secondgreen}0.2639 & \cellcolor{secondgreen}0.4283 & \cellcolor{bestred}0.5471 & \cellcolor{secondgreen}0.3433 & 0.4610 & \cellcolor{secondgreen}0.3901 & \cellcolor{secondgreen}0.5322 \\
 & \textbf{AMG-Fuse} & \textbf{--} &  & \cellcolor{bestred}0.3096 & \cellcolor{bestred}0.4376 & \cellcolor{secondgreen}0.5438 & \cellcolor{bestred}0.3445 & \cellcolor{bestred}0.4864 & \cellcolor{bestred}0.4083 & \cellcolor{bestred}0.5414 \\ \hline
\end{tabular}
\end{adjustbox}
\end{table}

\begin{table}[t]
\centering
\caption{Quantitative comparison of all methods in real-world scenarios. The marked \colorbox{bestred}{red} indicates the best score and the marked \colorbox{secondgreen}{green} indicates the second-best score.}
\label{tabreal}
\renewcommand{\arraystretch}{1}
\setlength{\tabcolsep}{6pt}      
\begin{adjustbox}{width=\textwidth}
\begin{tabular}{l|c|ccccccc} 
\hline
\textbf{Methods} & \textbf{Pub.} & $Q_{MI} \uparrow$ & $Q_{G} \uparrow$ & $Q_{M} \uparrow$ & $VIF \uparrow$ & $Q_{CB} \uparrow$ & $SSIM \uparrow$ & $Q^{AB/F} \uparrow$ \\ \hline
LRRNet & PAMI 23 & 0.5429 & 0.3944 & 0.9170 & \cellcolor{secondgreen}0.4079 & 0.4694 & 0.4174 & 0.4773 \\
Text-DiFuse & NeurIPS 24 & 0.5174 & 0.3325 & 0.9384 & 0.3850 & 0.4108 & 0.3848 & 0.4657 \\
EMMA & CVPR 24 & \cellcolor{secondgreen}0.5497 & 0.3884 & 0.8056 & 0.3264 & 0.4470 & 0.4342 & 0.5301 \\
Text-IF & CVPR 24 & 0.5258 & \cellcolor{bestred}0.5748 & \cellcolor{secondgreen}1.0852 & 0.3566 & \cellcolor{bestred}0.4936 & 0.4972 & \cellcolor{bestred}0.6800 \\
GIFNet & CVPR 25 & 0.5085 & 0.3673 & 0.8833 & 0.3021 & 0.4250 & 0.4489 & 0.4605 \\
SAGE & CVPR 25 & 0.5200 & 0.4786 & 1.0538 & 0.3647 & 0.4298 & 0.4816 & 0.5314 \\
AWFusion & INFFus 26 & 0.5032 & 0.4547 & 1.0028 & 0.3523 & 0.4649 & \cellcolor{secondgreen}0.5036 & 0.5860 \\ \hline
\textbf{AMG-Fuse} & \textbf{--} & \cellcolor{bestred}0.7234 & \cellcolor{secondgreen}0.5114 & \cellcolor{bestred}1.1022 & \cellcolor{bestred}0.4373 & \cellcolor{secondgreen}0.4786 & \cellcolor{bestred}0.5202 & \cellcolor{secondgreen}0.5957 \\ \hline
\end{tabular}
\end{adjustbox}
\end{table}

\subsection{Qualitative Comparison}
Figure~\ref{fig6} presents fusion results under three types of adverse weather conditions.  For most fusion methods, the visible input is first restored by AdaIR before fusion. AMG-Fuse (In Clean) represents the fusion result obtained using clean source images as input, serving as an upper-bound reference.

\textbf{Snow Weather:}
While AdaIR effectively removes snow interference, it causes significant detail loss, resulting in blurred output for LRRNet, Text-IF, and SAGE. In contrast, our method (AMG-Fuse) effectively eliminates degradation artifacts, preserves key details, and fully exploits the complementary information between modalities.

\textbf{Rain Weather:}
AdaIR removes rain streaks but oversmooths textures, leading to diminished scene details.  LRRNet and GIFNet fail to sufficiently highlight infrared cues, while EMMA, Text-IF, and SAGE enhance occluded visible targets but fail to capture fine infrared textures. AMG-Fuse, however, maintains a strong balance between multi-modal interaction and subtle visible details.

\textbf{Haze Weather:}
Haze introduces depth-dependent degradation, complicating feature extraction. LRRNet, Text-DiFuse, and SAGE exhibit significant contrast loss, while EMMA and Text-IF produce outputs that diverge from perceptual expectations. AWFusion introduces color distortions, further degrading visual fidelity. In contrast, AMG-Fuse effectively suppresses haze artifacts, generating visually natural and structurally rich fused images.

\textbf{Fusion in Clean Scene}
As shown in Figure~\ref{fig6}, we compare AMG-Fuse with three representative fusion methods in clean environments. Despite being designed for degraded scenarios, AMG-Fuse demonstrates robust multi-modality feature extraction and produces competitive results in ideal conditions, showcasing its adaptability and generalization capability.

\begin{figure}[t]
  \centering
   \includegraphics[width=1\linewidth]{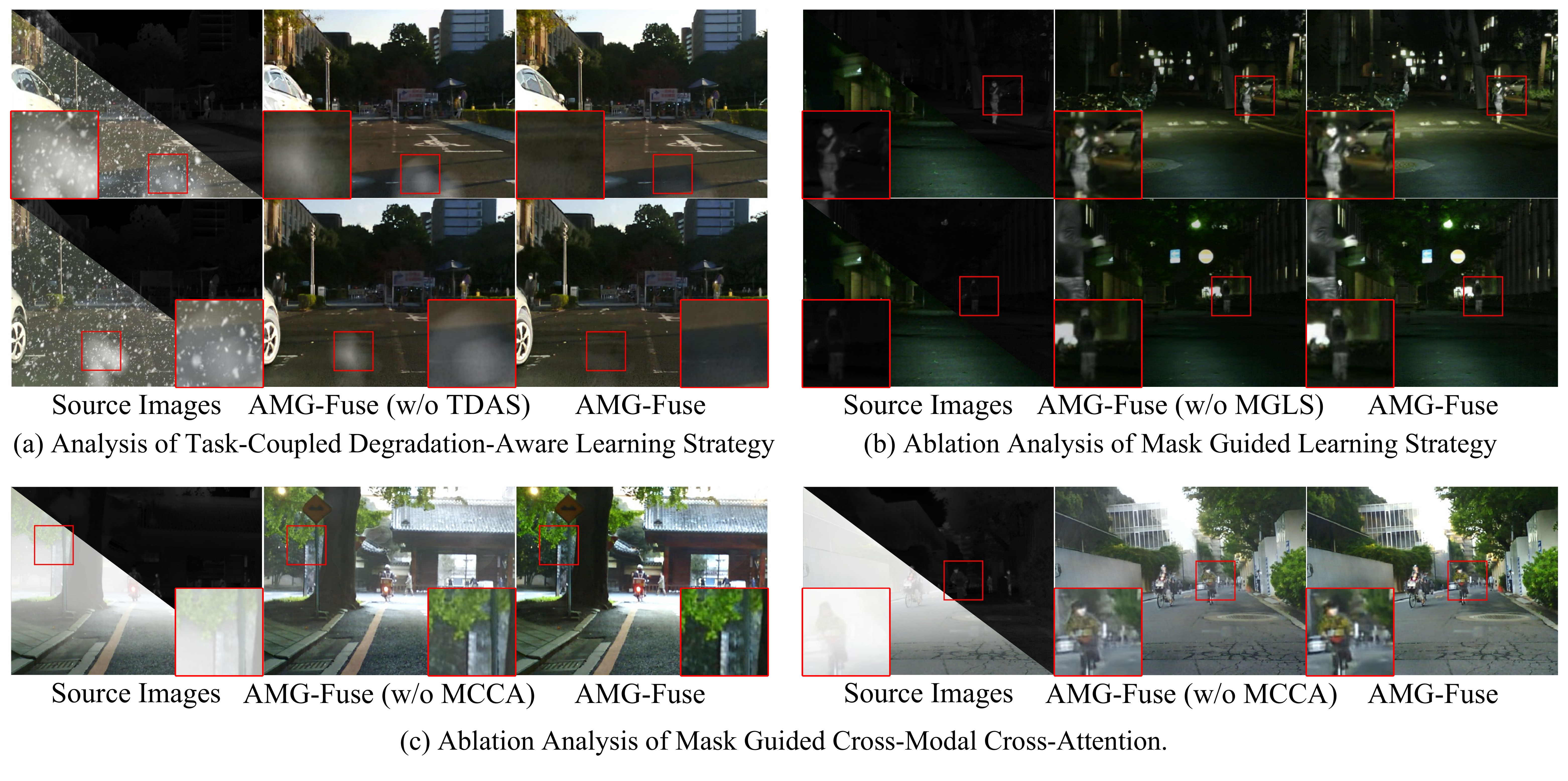}
  \caption{Visual analysis of ablation studies: (a) TDAS component ablation; (b) MGLS component ablation; (c) MCCA component ablation.}
   \label{fig7}
\end{figure}

\subsection{Quantitative Comparison}
Table~\ref{tab1} presents the quantitative results of AMG-Fuse and competing methods under the three adverse weather conditions. AMG-Fuse consistently ranks in the top two across all metrics and scenarios, outperforming other methods in several key metrics. Specifically, AMG-Fuse improves 3.67\%, 3.86\% and 3.56\% over the sub-optimal method in snow, rain, and haze scenes, respectively. 
To further validate the applicability of the proposed algorithm on real-world data, we conducted experiments on the real haze dataset, with quantitative results shown in Table~\ref{tabreal}. The proposed method achieves top-two scores across all metrics, demonstrating the effectiveness of the mask-guided learning strategy.

\begin{table}[t]
\centering
\caption{Ablation studies of the proposed components in three adverse weather scenarios. The cells highlighted in \colorbox{bestred}{red} denote the best results.}
\label{tab4}
\scriptsize
\renewcommand{\arraystretch}{1}
\setlength{\tabcolsep}{4pt}
\begin{tabular}{c|l|ccccccc}
\hline
\textbf{Source} & \textbf{Methods} & $Q_{MI} \uparrow$ & $Q_{G} \uparrow$ & $Q_{M} \uparrow$ & $VIF \uparrow$ & $Q_{CB} \uparrow$ & $SSIM \uparrow$ & $Q^{AB/F} \uparrow$ \\
\hline

\multirow{2}{*}{Snow}
& w/o TDAS            & 0.4620 & 0.3951 & 0.5097 & 0.3268 & 0.4878 & 0.3961 & 0.5012 \\
& \textbf{AMG-Fuse}   & \cellcolor{bestred}0.4960 & \cellcolor{bestred}0.4240 & \cellcolor{bestred}0.6408 & \cellcolor{bestred}0.3463 & \cellcolor{bestred}0.5062 & \cellcolor{bestred}0.4302 & \cellcolor{bestred}0.5519 \\
\hline

\multirow{2}{*}{Rain}
& w/o MGLS            & 0.3919 & 0.3910 & 0.6782 & 0.3531 & 0.4762 & 0.3928 & 0.4982 \\
& \textbf{AMG-Fuse}   & \cellcolor{bestred}0.4084 & \cellcolor{bestred}0.4163 & \cellcolor{bestred}0.6940 & \cellcolor{bestred}0.3582 & \cellcolor{bestred}0.4879 & \cellcolor{bestred}0.4080 & \cellcolor{bestred}0.5184 \\
\hline

\multirow{2}{*}{Haze}
& w/o MCCA            & 0.2803 & 0.4178 & 0.5012 & 0.3184 & 0.4587 & 0.3616 & 0.5217 \\
& \textbf{AMG-Fuse}   & \cellcolor{bestred}0.3096 & \cellcolor{bestred}0.4376 & \cellcolor{bestred}0.5438 & \cellcolor{bestred}0.3445 & \cellcolor{bestred}0.4864 & \cellcolor{bestred}0.4083 & \cellcolor{bestred}0.5414 \\
\hline

\end{tabular}
\end{table}

\subsection{Ablation Studies}

\textbf{Analysis of Task-Coupled Degradation-Aware Learning Strategy:}
As shown in Figure~\ref{fig7}(a) and Table~\ref{tab4}, removing TDAS (AMG-Fuse w/o TDAS) clearly degrades the fusion performance, especially under severe degradation where structural details are difficult to recover. Without the task-coupled degradation-aware constraint, the network relies mainly on multimodal feature extraction and lacks explicit guidance from the restoration model to preserve clean image characteristics. Consequently, the fused results show weakened structures and lower perceptual quality. The decline across all metrics further demonstrates the importance of TDAS in improving robustness against degraded visible inputs.

\textbf{Analysis of Mask Guided Learning Strategy:}
As shown in Figure~\ref{fig7}(b), removing MGLS (AMG-Fuse w/o MGLS) makes the fusion network depend more heavily on the static pixel distribution of the ``Pseudo Ground Truth''. Without mask-guided learning, the model lacks sufficient guidance for adaptive modality assignment in degraded regions, resulting in reduced flexibility in complex scenes. The quantitative results in Table~\ref{tab4} further support this observation. Although the drops in human-vision-related metrics such as $VIF$ and $Q_{CB}$ are relatively moderate, structure-sensitive metrics such as $Q_G$ and $Q_M$ decrease more obviously, with an average metric drop of 3.34\%.

\textbf{Analysis of Mask Guided Cross-Modal Cross-Attention:}
As shown in Figure~\ref{fig7}(c), removing MCCA (AMG-Fuse w/o MCCA) weakens cross-modal interaction and turns the network into a more unidirectional recovery structure. This limits the complementary information exchange between infrared and visible modalities, leading to inferior detail restoration and degraded-region recovery. Consistently, Table~\ref{tab4} shows that all metrics decrease after removing MCCA, with an average performance drop of about 6.9\%. These results verify that MCCA is essential for enhancing multi-modal synergy and improving the overall fusion quality of AMG-Fuse.

\begin{table}[t]
\centering
\caption{Detection accuracy (mAP) comparison on the M3FD dataset. The marked \colorbox{bestred}{red} indicates the best score and the marked \colorbox{secondgreen}{green} indicates the second-best score.}
\label{tab2}
\renewcommand{\arraystretch}{1} 
\setlength{\tabcolsep}{5pt}      
\begin{adjustbox}{width=\textwidth}
\begin{tabular}{l|c|cccccccc} 
\hline
\textbf{Methods} & \textbf{Pub.} & \textbf{People} & \textbf{Car} & \textbf{Bus} & \textbf{Lamp} & \textbf{Motor} & \textbf{Truck} & \textbf{mAP@0.5} & \textbf{mAP@0.5:0.95} \\ \hline
LRRNet & PAMI 23 & 0.785 & 0.909 & 0.912 & 0.764 & 0.721 & 0.808 & 0.817 & 0.484 \\
Text-DiFuse & NeurIPS 24 & 0.790 & 0.910 & 0.927 & 0.793 & 0.690 & 0.795 & 0.817 & 0.520 \\
EMMA & CVPR 24 & 0.784 & 0.904 & 0.890 & 0.722 & 0.690 & 0.780 & 0.795 & 0.501 \\
Text-IF & CVPR 24 & 0.812 & 0.920 & 0.923 & 0.806 & \cellcolor{bestred}0.744 & 0.811 & \cellcolor{secondgreen}0.836 & \cellcolor{secondgreen}0.535 \\
GIFNet & CVPR 25 & 0.806 & 0.906 & 0.919 & 0.777 & 0.671 & 0.779 & 0.810 & 0.506 \\
SAGE & CVPR 25 & \cellcolor{secondgreen}0.814 & 0.919 & 0.921 & \cellcolor{secondgreen}0.815 & 0.698 & \cellcolor{bestred}0.830 & 0.833 & 0.534 \\
AWFusion & INFFus 26 & 0.791 & \cellcolor{secondgreen}0.922 & \cellcolor{secondgreen}0.931 & 0.804 & 0.726 & \cellcolor{secondgreen}0.821 & 0.832 & 0.534 \\ \hline
\textbf{AMG-Fuse} & \textbf{--} & \cellcolor{bestred}0.837 & \cellcolor{bestred}0.931 & \cellcolor{bestred}0.934 & \cellcolor{bestred}0.816 & \cellcolor{secondgreen}0.728 & 0.813 & \cellcolor{bestred}0.843 & \cellcolor{bestred}0.541 \\ \hline
\end{tabular}
\end{adjustbox}
\end{table}
\begin{figure}[t]
  \centering
   \includegraphics[width=1\linewidth]{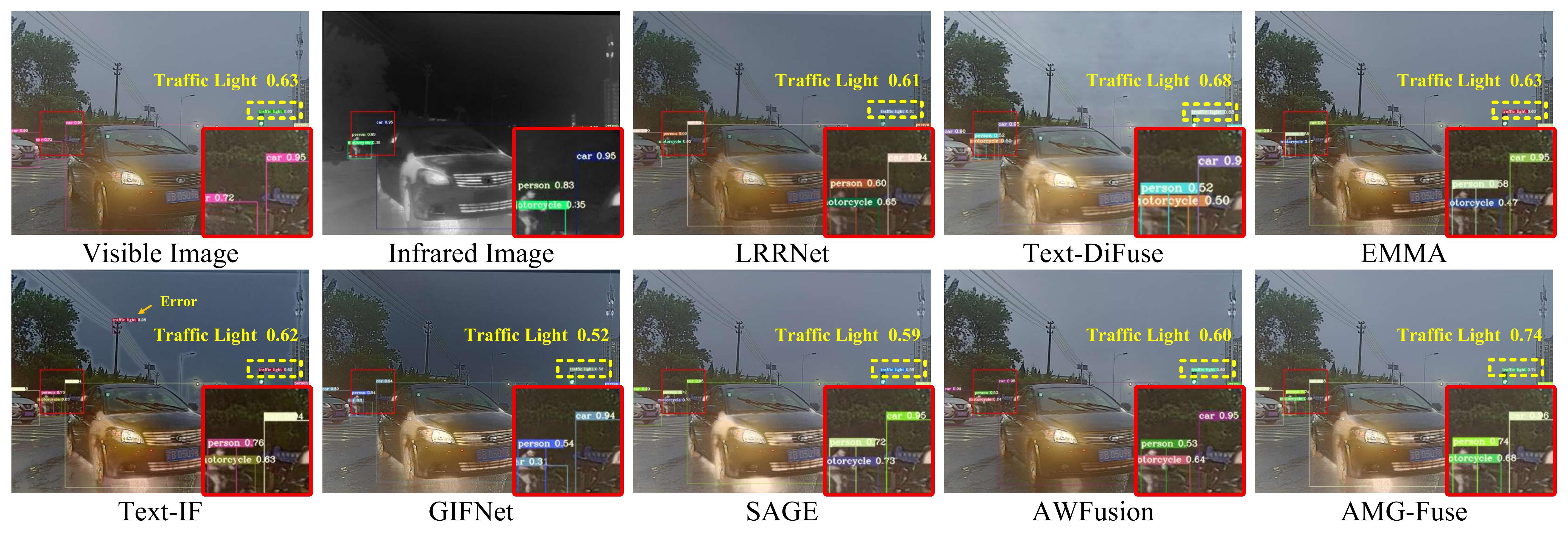}
   \caption{Qualitative comparison of all methods on object detection tasks.}
   \label{figObj}
\end{figure}

\subsection{Downstream Task Experiment}
To assess the practicality of AMG-Fuse in downstream tasks, we conduct object detection on the M3FD dataset using YOLOv7 \cite{r6}. As shown in Table~\ref{tab2}, our method ranks among the top two in five of six categories and achieves the best mAP@0.5 and mAP@[0.5:0.95]. Figure \ref{figObj} further demonstrates that our results offer higher accuracy in object detection. Overall, AMG-Fuse improves the usefulness of fused images for downstream tasks.

\begin{table}[t]
\centering
\caption{Quantitative comparison between the proposed method and the ``Pseudo Ground Truth'' generator EMMA on three datasets.}
\label{tabWGT}
\renewcommand{\arraystretch}{1} 
\setlength{\tabcolsep}{6pt}     
\begin{adjustbox}{width=\textwidth}
\begin{tabular}{l|l|c|ccccccc} 
\hline
\textbf{Source} & \textbf{Methods} & \textbf{Pub.} & $Q_{MI} \uparrow$ & $Q_{G} \uparrow$ & $Q_{M} \uparrow$ & $VIF \uparrow$ & $Q_{CB} \uparrow$ & $SSIM \uparrow$ & $Q^{AB/F} \uparrow$ \\ \hline
\multirow{2}{*}{MSRS} & EMMA & CVPR 24 & 0.6697 & 0.5624 & 0.7291 & 0.4165 & 0.5394 & 0.4746 & 0.6500 \\
 & \textbf{AMG-Fuse} & \textbf{--} & \cellcolor{bestred}0.7594 & \cellcolor{bestred}0.5952 & \cellcolor{bestred}1.5051 & \cellcolor{bestred}0.4706 & \cellcolor{bestred}0.5809 & \cellcolor{bestred}0.5044 & \cellcolor{bestred}0.6749 \\ \hline
\multirow{2}{*}{M3FD} & EMMA & CVPR 24 & 0.5646 & 0.4827 & 0.5595 & 0.3829 & 0.4880 & 0.4787 & 0.6032 \\
 & \textbf{AMG-Fuse} & \textbf{--} & \cellcolor{bestred}0.7010 & \cellcolor{bestred}0.5376 & \cellcolor{bestred}1.2994 & \cellcolor{bestred}0.4713 & \cellcolor{bestred}0.4979 & \cellcolor{bestred}0.5069 & \cellcolor{bestred}0.6384 \\ \hline
\multirow{2}{*}{LLVIP} & EMMA & CVPR 24 & 0.4437 & 0.5005 & 0.2314 & 0.3301 & \cellcolor{bestred}0.4559 & 0.4339 & 0.5969 \\
 & \textbf{AMG-Fuse} & \textbf{--} & \cellcolor{bestred}0.4684 & \cellcolor{bestred}0.6228 & \cellcolor{bestred}0.7938 & \cellcolor{bestred}0.3952 & 0.4479 & \cellcolor{bestred}0.4574 & \cellcolor{bestred}0.6907 \\ \hline
\end{tabular}
\end{adjustbox}
\end{table}

\subsection{The Discussion and Generation of ``Pseudo Ground Truth''}
\label{sec:WGT}

``Pseudo Ground Truth'' is introduced as a guidance signal to help the model learn modality allocation. Two concerns naturally arise: (1) whether it may limit the performance upper bound of the proposed method, and (2) whether it may introduce modality bias into the fusion process. To clarify these issues, we provide the following analysis:

\begin{itemize}
    \item The proposed method does not depend on ``Pseudo Ground Truth'' alone. It is used as a soft guidance signal, while clean source images provide the main supervision for learning modality allocation.

    \item ``Pseudo Ground Truth'' is generated from clean images and thus preserves more reliable structure and texture than weather-degraded inputs. Even with potential bias, it mainly helps stabilize early training rather than limiting the final performance.

    \item The loss weight of MGLS decays over training: it is larger early on to improve stability and convergence, and becomes smaller later so the model relies more on clean-image supervision. This makes ``Pseudo Ground Truth'' an early guide without constraining later learning.
\end{itemize}

To empirically verify that ``Pseudo Ground Truth'' does not bound our model, we generate pseudo targets with EMMA from clean multi-modal images and compare EMMA with AMG-Fuse under the same inputs. If EMMA-derived guidance were an upper bound, EMMA would outperform our model. However, Table~\ref{tabWGT} shows that AMG-Fuse outperforms EMMA on most metrics across three standard datasets, demonstrating that our model is not constrained by pseudo targets and learns stronger overall fusion representations.
In summary, ``Pseudo Ground Truth'' is mainly used to stabilize early training and guide optimization. EMMA is adopted to provide a reliable prior for faster convergence rather than an optimal reference. We fine-tune EMMA on the clean subset of AWMM-100k (1,000 images), initialized from its original pretrained weights.

\subsection{Limitation and Future Work}
Our experiments show that the Histogram Transformer in the Mask-Guided Feature Extraction Module increases computational cost. For an input of $224 \times 224$, our method requires $242.03$G FLOPs and $59.74$M parameters. In future work, we will improve efficiency to better support real-world applications.

\section{Conclusion}

In this paper, we present AMG-Fuse, a mask-guided multi-modality image fusion framework for both adverse weather and clean conditions. To ease optimization while avoiding overfitting to pseudo targets, we introduce ``Pseudo Ground Truth'' and derive a modality-contribution mask from the mapping between source images and fused outputs. This mask guides cross-modal interaction, while the mask-guided and task-coupled degradation-aware learning strategies jointly enhance feature restoration and modality interaction under complex degradation. Extensive experiments demonstrate that AMG-Fuse consistently outperforms state-of-the-art methods across diverse scenarios.

\section*{Acknowledgement}

This work was supported in part by the Basic and Applied Basic Research of Guangdong Province under Grant
2023A1515140077, in part by the Natural Science Foundation of Guangdong Province under Grant 2024A1515011880, in part by the National Natural Science Foundation of China under Grant 52374166 and 62201149, in part by the Research Fund of Guangdong-Hong Kong-Macao Joint Laboratory for Intelligent MicroNano Optoelectronic Technology under Grant 2020B1212030010, and in part by the Yunnan Fundamental Research Projects under Grant 202301AV070004 and Grant 202501AS070123.

\bibliographystyle{splncs04}
\bibliography{main}
\end{document}